\documentclass{article}

\usepackage[final]{neurips_2022_ml4ps}

\usepackage[utf8]{inputenc} 
\usepackage[T1]{fontenc}    
\usepackage{hyperref}       
\usepackage{url}            
\usepackage{booktabs}       
\usepackage{amsfonts}       
\usepackage{nicefrac}       
\usepackage{microtype}      
\usepackage{xcolor}         

\usepackage{amsmath}
\usepackage{mathtools}

\usepackage{subfigure}
\usepackage{varwidth}
\usepackage{graphicx}
\usepackage{wrapfig}
\usepackage{comment}

\title{A Curriculum-Training-Based Strategy for Distributing Collocation Points during Physics-Informed Neural Network Training}

\author{%
  \href{http://orcid.org/0000-0002-8712-4035}{Marcus Münzer}\thanks{Alternative Email: \texttt{marcus.muenzer@mkl-gmbh.de}} \\
  Institute of Informatics \\
  Ludwig Maximilians University Munich\\
  Munich, D-80538, Germany \\
  \texttt{m.muenzer@campus.lmu.de} \\
  \And
  \href{https://orcid.org/0000-0002-5926-0566}{Chris Bard}\\
  Geospace Physics Lab \\
  NASA Goddard Space Flight Center \\
  Greenbelt, MD, United States \\
  \texttt{christopher.bard@nasa.gov} \\
}

\begin{document}

\maketitle

\begin{abstract}
Physics-informed Neural Networks (PINNs) often have, in their loss functions, terms based on physical equations and derivatives. In order to evaluate these terms, the output solution is sampled using a distribution of collocation points. However, density-based strategies, in which the number of collocation points over the domain increases throughout the training period, do not scale well to multiple spatial dimensions. To remedy this issue, we present here a curriculum-training-based method for lightweight collocation point distributions during network training. We apply this method to a PINN which recovers a full two-dimensional magnetohydrodynamic (MHD) solution from a partial sample taken from a baseline MHD simulation. We find that the curriculum collocation point strategy leads to a significant decrease in training time and simultaneously enhances the quality of the reconstructed solution.
\end{abstract}

\section{Introduction}\label{intro}

Solving partial differential equations (PDEs) is a fundamental task in space physics. Although numerical algorithms have tremendously improved over the years [\cite{wein21}], providing solutions over a full space-time domain often requires large-scale, computationally-intensive simulations. As an alternative to such simulations, machine learning (ML) techniques have become popular in recent years, fusing physics and data (e.g., soft proton predictions [\cite{muenz21}], solar flare forecasts [\cite{nish21}], or space weather forecasts [\cite{campo19}]). However, most of these approaches rely on huge datasets which appropriately cover the relevant domain. This approach cannot work for domains with a sparsity problem, e.g., spacecraft missions in planetary magnetospheres in which the amount of data recorded is very tiny compared to the amount of data inherent in the global system.

One way to address this sparsity problem is to numerically solve the physical equations over the full domain given specific initial and boundary conditions. However, such global simulations can be quite costly, requiring many thousands of computer core hours. Additionally, they are not yet sophisticated enough to assimilate data and modify their solutions appropriately to match observations. As an alternative, PINNs can alleviate the sparsity problem by combining data observations and physical constraints in their loss functions to reconstruct global contexts around local data samples (e.g. the one-dimensional magnetohydrodynamics reconstructions in [\cite{bard21}]). This has the potential to provide a lightweight alternative to global simulations, but only if the PINN is able to train quickly.

In order to generate physically-valid solutions, PINNs ensure that their output solutions correctly correspond to their inherent physical equations [\cite{rai19, karn21}]. These calculations require so-called "collocation points" distributed through the domain at which the solution and its derivatives are evaluated and tested against the physical equations. The simplest strategy for distributing collocation points is to randomly scatter them throughout the domain and increase the density of points as network training progresses. Other strategies include evolutionary sampling, which gradationally aggregates points in areas of high PDE loss residuals [\cite{daw22}], or the importance method, which samples collocation points according to a distribution proportional to the loss function [\cite{Nab21}]. 
Currently, one of the major obstacles in developing a PINN-based solver for three-dimensional reconstruction of space plasmas is the poor scaling to multiple dimensions. The previously mentioned density-based strategy requires increasingly significant computational resources as the dimensionality goes up [\cite{bard21}]. In this paper, we present two novel curriculum-training-based approaches using lightweight collocation point distributions that better scale to multiple dimensions. We apply these methods to a PINN which recovers a full two-dimensional MHD solution from a partial sample of simulated spacecraft observations. We find that the new strategy leads to a significant decrease in training time and simultaneously enhances the quality of the reconstructed solution. 

\section{Model Setup}\label{setup}
We use a PINN designed to reconstruct two-dimensional MHD solutions given partial linear samples of the original solution, mimicking the reconstruction of space environments from satellite data. This extends the one-dimensional network of [\cite{bard21}] to two dimensions.
The rebuild of higher-dimensional plasma space-time is interpreted as a regression problem of form
\begin{align}
\mathbf{U}^{st} \approx f([\mathbf{x}; \mathbf{y}; \mathbf{t}] \mid \mathbf{\theta}) \coloneqq \mathbf{U}_{net}(x, y, t)
\end{align}
where $\mathbf{U}^{st}$ is the full MHD solution, $\mathbf{x}$ and $\mathbf{y}$ are the spatial positions, and $\mathbf{t}$ is time.
$[~;~]$ denotes concatenation and $\theta$ are model parameters that are to be found through training. 
The approximated solution $\mathbf{U}_{net}$ follows both a physical constraint
\begin{align}
    \frac{\partial \mathbf{U}_{net}}{\partial t} + F(\mathbf{U}_{net}, \frac{\partial\mathbf{U}_{net}}{\partial x}) + G(\mathbf{U}_{net}, \frac{\partial\mathbf{U}_{net}}{\partial y})= 0
\end{align}
and a boundary data constraint
\begin{align}
\mathbf{U}_{net}(x_i, y_i, t_i) = \mathbf{U}^{st}_i, \; i = 1 \dots N_d
\end{align}
for $N_d$ coordinates of $(x, y, t)$ in space-time. $F$ and $G$ are the two-dimensional MHD fluxes (see Appendix \ref{mhd} for further details).

In order to mimic satellite observations in space, every experiment builds on four randomly chosen linear trajectories that are constrained to cover the full time range but only a small part of the spatial domain. We emulate sparse craft measurements by sampling a portion of approximately $10^{-6}$ points per trajectory in comparison to the full space-time grid.
As recent PINNs primarily incorporate dense neural networks [\cite{chi22}, e.g. \cite{henn20}], we implement our PINN as a Multi-Layer Perceptron Regressor [\cite{ros58, coll43}]. Its weights are trained by the ADAM optimizer [\cite{king14}] while respecting a predefined learning rate schedule [\cite{you19}] to improve the convergence curve of the model [\cite{bard21}]. Hyperparameters are tuned using the Tree-structured Parzen Estimator Approach [\cite{berg11}] and trials get pruned by applying median pruning. We enforce five startup runs during which the pruning is disabled.

The loss function is defined as $L_{PINN} = (L_{data} + \lambda L_{phys})/(1 + \lambda)$ where $L_{data}, L_{phys}$ represent the data and physical loss, and $\lambda$ is a trade-off parameter balancing these two. The physical loss is evaluated at collocation points distributed within the reconstruction domain. Previous work reconstructing plasma solutions around spacecraft trajectories [\cite{bard21}] used density-based collocation point sampling in which the density of collocation points in space-time increases over the training interval. However, this does not scale well to higher spatial dimensions; in fact, the density-based run time can be expressed as $\mathcal{O}(n_T + n_C^{d + 1})$ [\cite{mal22}], where $n_T$ is the number of data points and $n_C$ is the number of collocation points required to satisfy a specific collocation density in each of the $d$ spatial dimensions and one time dimension. Increasing the density has a larger effect on runtime with increasing dimensionality.

Hence, the collocation point sampling method needs to be independent of the problem dimensionality for the PINN to scale well. Thus, we define the set of sampled collocation points per epoch as a constant matching the number of spacecraft observations, which results in a linear runtime of $\mathcal{O}(n_T + n_C) \cong \mathcal{O}(n_T)$. This provides a significant speedup in comparison to density-based collocation point sampling, but we must be intentional in how we distribute the points.

\section{Curriculum Strategies}
The beginnings of curriculum learning in the field of ML can at least be backtraced to [\cite{elm93}]. Inspired by human education, it is based on the idea of guiding a learning process by increasing the difficulty level of tasks over time [\cite{ben09, sov21}]. Originally, this is done by feeding a model easy-to-learn data first, and then gradually transitioning to more complex data concepts [\cite{ben09}]. Curriculum training has already demonstrated its potential to improve various machine learning models for different applications, for instance MLPs for healthcare prediction [\cite{bouri20}], CNNs for computer vision [\cite{born21}], RNNs for language modeling [\cite{shi14}], and even PINNs for solving one-dimensional MHD problems [\cite{krish21}]. Here, we propose two novel curriculum strategies for our PINN training.

\begin{figure}[!h]
\centering
\fbox{
\begin{varwidth}{\linewidth}
\subfigure[Cuboid sampling step 0]{\includegraphics[width=4cm]{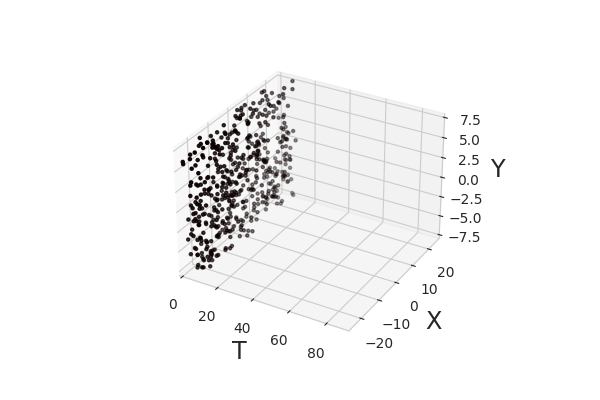}}
\subfigure[Cuboid sampling step 1]{\includegraphics[width=4cm]{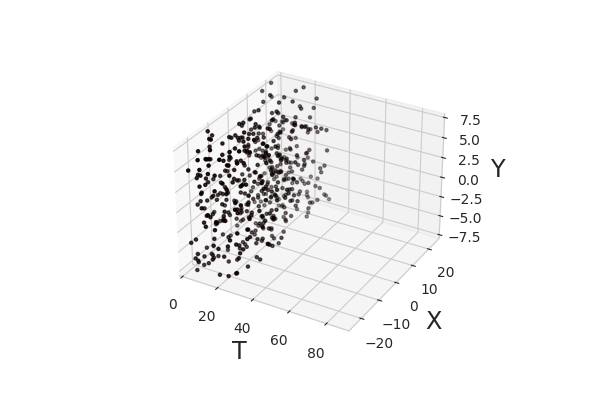}}
\subfigure[Cuboid sampling step 2]{\includegraphics[width=4cm]{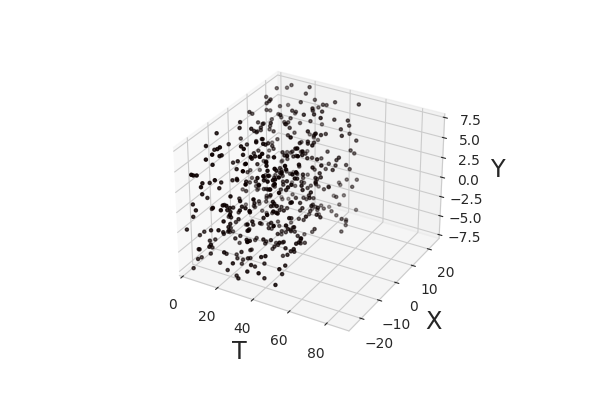}}\\
\subfigure[Cylinder sampling step 0]{\includegraphics[width=4cm]{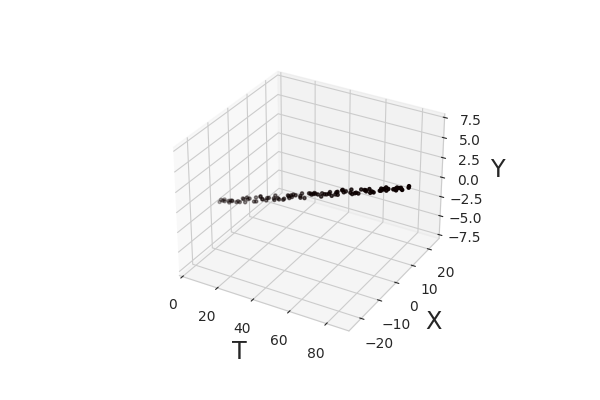}}
\subfigure[Cylinder sampling step 1]{\includegraphics[width=4cm]{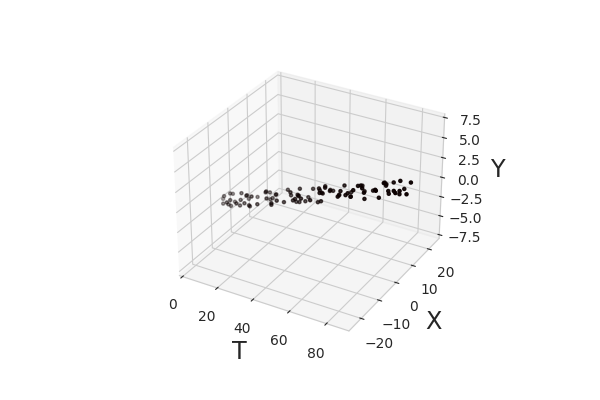}}
\subfigure[Cylinder sampling step 2]{\includegraphics[width=4cm]{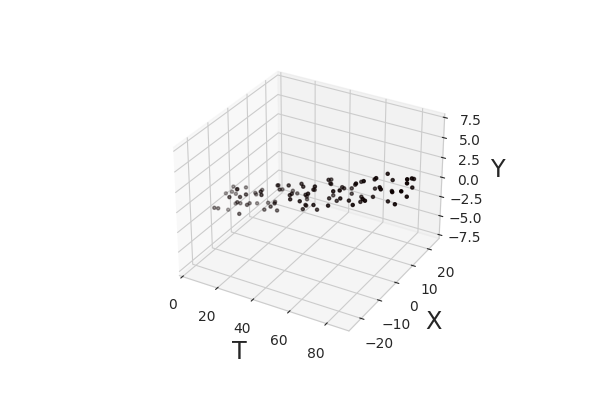}}
\end{varwidth}}
\caption{Evolution of collocation point distributions for the Cuboid and Cylinder Curriculum methods over several training epochs.} 
\label{fig_curr}
\end{figure}

First, we define a sampling method whereby the initial collocation point distribution is a cuboid covering the full space domain but only in a very small time range starting at $t_0$. As training progresses, the cuboid gradually extends along the time axis, eventually encompassing the full reconstruction domain. In this way, the model first builds up an accurate reconstruction of the time-dependent behavior at initial times and then expands this knowledge throughout the training (an example is shown in Figure \ref{fig_curr}).
Analogously, the model could also be supported in its training by stretching out the cuboid to the full (x, t) domain and extending along y, or to the full (y, t) domain and extending along x. This is referred to as the \textit{cuboid curriculum method}.

Alternatively, we hypothesize that regions closer to the sparse training data points are easier to reconstruct than ones further away in space-time. To take advantage of this, we distribute collocation points in small-radius spheres around the labeled training data and then stepwise increase the radii of these bubbles until they cover the whole space-time domain. For our specific use-case here (Sec. \ref{setup}), the data points are arranged linearly in space; consequently, the collocation points are located in a cylinder with increasing radius (e.g. Figure \ref{fig_curr}) around the linear spacecraft trajectories. This strategy is referred to as the \textit{cylinder curriculum method}, though it can be generalized to a \textit{bubble method}.

\section{Results}\label{res}
We test the curriculum training methods on three different datasets.
First, an MHD reconnection benchmark from [\cite{birn01}] (name: GEM) of the following resolution: $-25.60 \leq x < 25.60$ with $\Delta x = 0.04$, $-7.68 \leq y < 7.68$ with $\Delta y = 0.04$, and $0 \leq t \leq 90$ with $\Delta t = 0.45$; resulting in a mesh of $1280 \times 384 \times 201$ (correspondingly $98,795,520 \approx 10^8$ points in total).
Second, we use a 2D Riemann Problem based on Case 3 from [\cite{lisk03}] (name: LW3). Since this is technically a hydrodynamic benchmark, we artificially inflate its plasma state vectors by adding the magnetic variables $B$ and setting them to zero. We run the base simulation until $t=0.35$ with an output every $\Delta t = 5 \times 10^{-3}$. The spatial dimensions are resolved as $0 \leq x < 1$ with $\Delta x \approx 5 \times 10^{-4}$, and $0 \leq y < 1$ with $\Delta y \approx 5 \times 10^{-4}$; leading to a grid of $2048 \times 2048 \times 71$ and a total of $297,795,584 \approx 3 \times 10^9$ voxels.
Finally, we explore an MHD vortex [\cite{OTvor79}] designed to study turbulence (name: OT).  It is represented as a mesh of $1024 \times 1024 \times 50$ (consequently $98,795,520 \approx 10^8$ overall points) where $0 \leq x < 1$ with $\Delta x \approx 10^{-2}$, $0 \leq y < 1$ with $\Delta y \approx 10^{-3}$, and $0 \leq t < 1$ with $\Delta t = 0.02$.\\

\setlength{\intextsep}{0.5pt}
\setlength{\columnsep}{19pt}
\begin{wrapfigure}{l}{8cm}
\centering
\fbox{
\includegraphics[width=8cm]{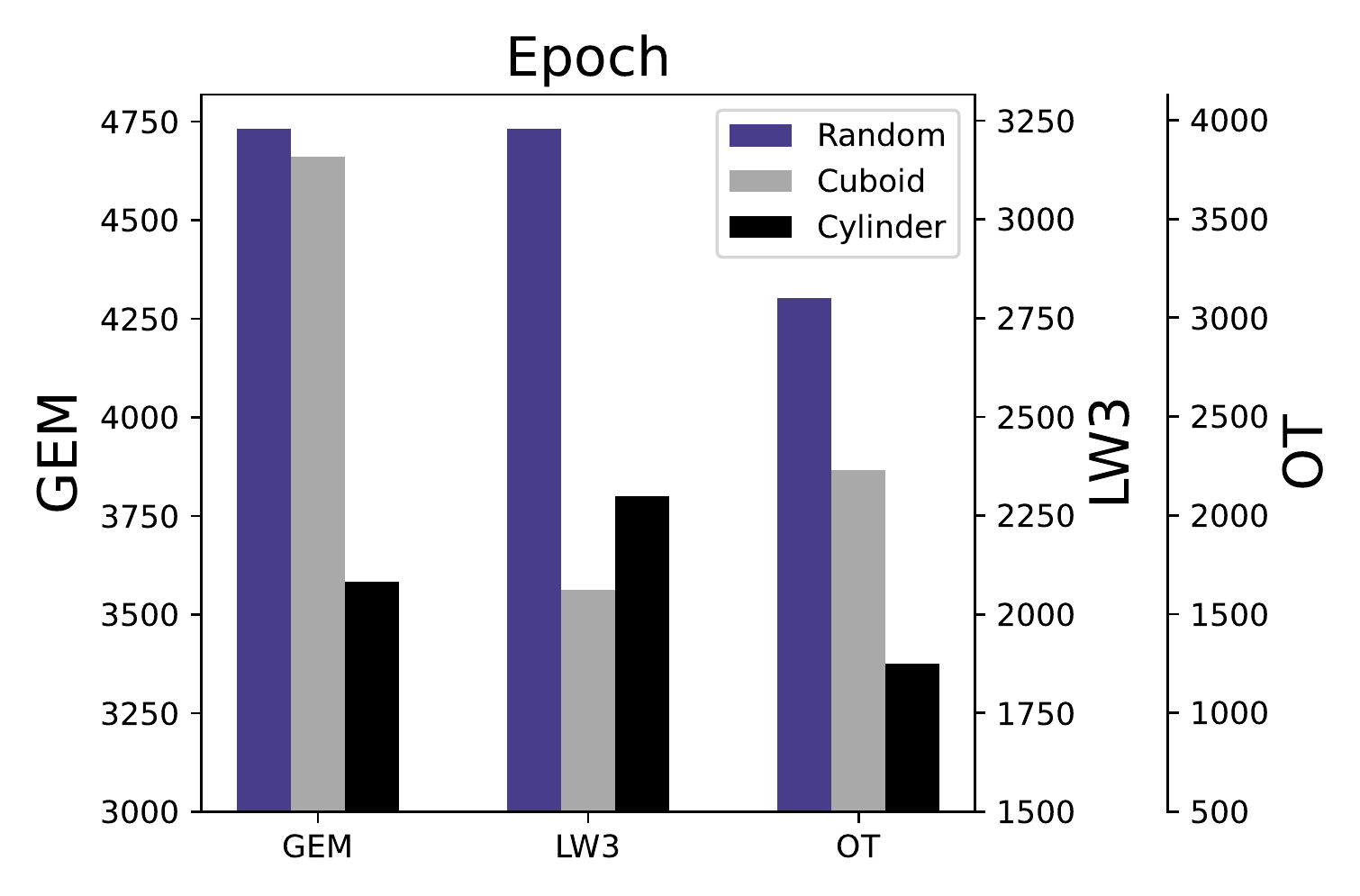}
}
\caption{PINN convergence epoch for the GEM data (left axes), the LW3 data (center axes), and the OT test data (right axes).} 
\label{epoch}
\vspace{6.5pt}
\end{wrapfigure}

\vspace{-10pt}
In addition to the cuboid and cylinder curriculum methods, we test a baseline strategy of randomly distributing the collocation points throughout the full space-time domain, also resampling them every epoch.
We train the models for a maximum of 5,000 epochs, with the curriculum training covering the initial $30\%$ (1500 epochs) of the training period. We extend the collocation point distribution five (15) times for the cuboid (cylinder) method, or every 300 (100) epochs until full coverage. This was experimentally chosen to ensure high accuracy.
50 experiments were run for each dataset and curriculum method. Each trial took about six minutes on an Nvidia Tesla V100 GPU at the Compute Cloud of the Leibniz Supercomputing Centre (45 computing hours total).

\begin{wrapfigure}{r}[-9.4pt]{8cm}
\centering
\raisebox{0pt}[\dimexpr\height-3.4\baselineskip\relax]{%
    \fbox{
    \includegraphics[width=8cm]{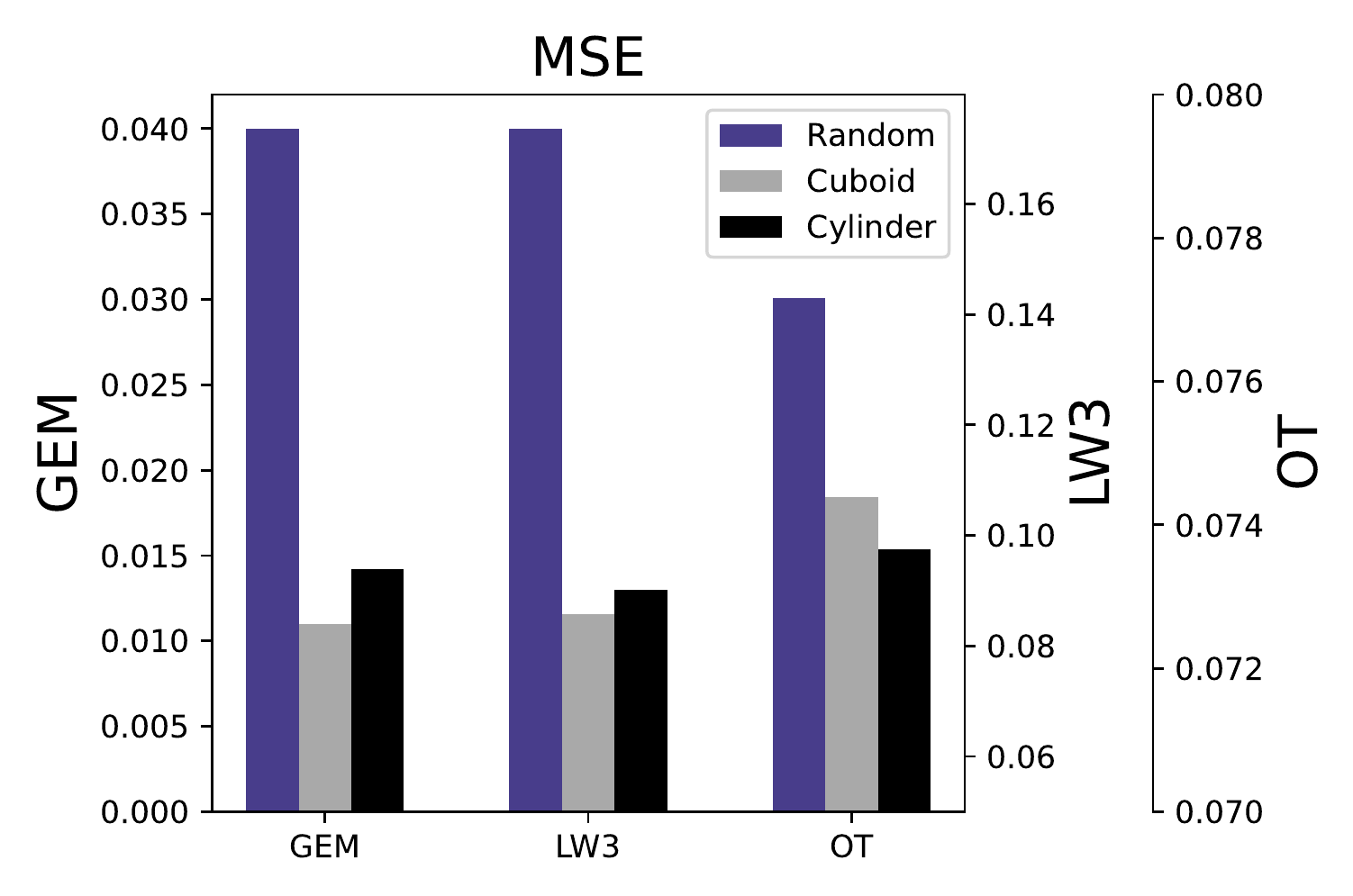}
    }
}
\caption{PINN mean squared error for the GEM data (left axes; purple), the LW3 data (center axes), and the OT test data (right axes).} 
\label{mse}
\vspace{-5pt}
\end{wrapfigure}

Figures \ref{epoch} and \ref{mse} show the mean squared error and the training speed as measured by the average number of epochs to convergence. We find that the curriculum-trained networks converged faster and to a lower error than the random-collocation-trained network.
On average, the curriculum networks took $\approx35\%$ fewer epochs to converge, though the cuboid+GEM model took approximately the same amount of epochs as the random+GEM model. Thus, the cylinder (or bubble) method performed slightly better overall than the cuboid method, albeit this depends on the scenario. Finally, the reconstruction accuracy increased: the cuboid and cylinder curriculum methods achieved an improvement of up to $\approx72\%$, and $\approx32\%$ on average.

\newpage

\setlength{\intextsep}{10pt}
\setlength{\columnsep}{10pt}

\section{Discussion and Conclusion}\label{disc}
We find that the curriculum training method for distributing collocation points significantly enhances PINN MHD reconstructions, simultaneously boosting accuracy and reducing time to convergence. However, we note that our results depend on the scenario and the models' initializations. Although the experiments here are robust, it is not guaranteed that our findings are universally applicable to every plasma environment. We note that hyperparameters, including the number of collocation points and the rate of expansion for the distribution, greatly influence the results. Future work may look for heuristics to fully exploit these strategies, possibly taking advantage of local loss errors [\cite{Nab21}]. Alternatively, the methods can be applied to 3D simulations. We envision an even more powerful runtime supremacy here, though this remains to be tested.

{
\small
\bibliography{bibliography}{}
}

\appendix

\section{Appendix}\label{app}

\subsection{Reproducing Experiments}

The experiments are based on the code of [\cite{muenz22repo}].\\
We use the following datasets: GEM [\cite{muenz22gem}], LW3 [\cite{muenz22lw3}], and OT [\cite{muenz22ot}].

\subsection{MHD Equations}\label{mhd}

The three-dimensional magnetohydrodynamic equations (using primitive variables) are:
\begin{align}
        \frac{\partial\rho}{\partial t} + \nabla\cdot(\rho \vec{v}) = 0\\
        \rho\frac{\partial\vec{v}}{\partial t} + \rho\vec{v}\cdot\nabla\vec{v} = -\nabla P + (\nabla\times\vec{B})\times\vec{B} + \textcolor{red}{\rho\nu\nabla^2\vec{v}}\\
        \frac{\partial P}{\partial t} + \vec{v}\cdot\nabla P + \gamma P \nabla\cdot\vec{v} = 0\\
        \frac{\partial \vec{B}}{\partial t} = \nabla\times\left(\vec{v}\times\vec{B}\right) + \textcolor{red}{\eta\nabla^2\vec{B}}\\
        \nabla\cdot\vec{B} = 0
\end{align}
We note that the ideal MHD equations do not contain the terms in red, which are the effects of viscosity ($\nu\nabla^2\vec{v}$) and resistivity ($\eta\nabla^2\vec{B}$). The simulations used to generate the test problems in this paper (LW3, GEM, OT) did not include these terms. However, based on results from [\cite{mich20}] and [\cite{bard21}], we add these terms to the PINN loss function in order to smooth out and better reconstruct the final solution near discontinuities (e.g., shock fronts). We clarify that the PINN cannot reproduce a discontinuous solution; however, adding viscosity and resistivity terms helps reproduce a steep, smooth solution where there is a discontinuity in the original simulation.

For the 2D PINN used in this paper, we convert these equations to two spatial dimensions by assuming $\partial/\partial z = \partial^2/\partial z^2 = 0$. The full set of equations, expanded out to individual derivatives, is:

\begin{subequations}
\begin{align*}
    \label{eq:rho}
    \frac{\partial\rho}{\partial t} + \frac{\partial (\rho v_x)}{\partial x}+\frac{\partial (\rho v_y)}{\partial y} = 0\\
    \rho\frac{\partial v_x}{\partial t} + \rho v_x \frac{\partial v_x}{\partial x} + \rho v_y \frac{\partial v_x}{\partial y} +\frac{\partial P}{\partial x} + B_z \frac{\partial B_z}{\partial x} + B_y\left(\frac{\partial B_y}{\partial x} - \frac{\partial B_x}{\partial y}\right) - \rho\nu\left(\frac{\partial^2 v_x}{\partial x^2} + \frac{\partial^2 v_x}{\partial y^2}\right) = 0\\ 
    \rho\frac{\partial v_y}{\partial t} + \rho v_x \frac{\partial v_y}{\partial x} + \rho v_y \frac{\partial v_y}{\partial y} + \frac{\partial P}{\partial y} - B_x\left(\frac{\partial B_y}{\partial x} - \frac{\partial B_x}{\partial y}\right) + B_z\frac{\partial B_z}{\partial y} - \rho\nu\left(\frac{\partial^2 v_y}{\partial x^2} + \frac{\partial^2 v_y}{\partial y^2}\right) = 0\\ 
    \rho\frac{\partial v_z}{\partial t} + \rho v_x \frac{\partial v_z}{\partial x} + \rho v_y \frac{\partial v_z}{\partial y} - B_y\frac{\partial B_z}{\partial y} - B_x\frac{\partial B_z}{\partial x} - \rho\nu\left(\frac{\partial^2 v_z}{\partial x^2} + \frac{\partial^2 v_z}{\partial y^2}\right) = 0\\
    \frac{\partial P}{\partial t} + v_x\frac{\partial P}{\partial x} + v_y\frac{\partial P}{\partial y} + \gamma P \left(\frac{\partial v_x}{\partial x}+\frac{\partial v_y}{\partial y}\right) = 0\\ 
    \frac{\partial B_x}{\partial t} - \frac{\partial}{\partial y}\left(v_x B_y - v_y B_x\right) - \eta\left(\frac{\partial^2 B_x}{\partial x^2} + \frac{\partial^2 B_x}{\partial y^2}\right) = 0  \\ 
    \frac{\partial B_y}{\partial t} + \frac{\partial}{\partial x}\left(v_x B_y - v_y B_x\right) - \eta\left(\frac{\partial^2 B_y}{\partial x^2} + \frac{\partial^2 B_y}{\partial y^2}\right) = 0\\ 
    \frac{\partial B_z}{\partial t} - \frac{\partial}{\partial x}\left(v_z B_x - v_x B_z\right) + \frac{\partial}{\partial y}\left(v_y B_z - v_z B_y\right) - \eta\left(\frac{\partial^2 B_z}{\partial x^2} + \frac{\partial^2 B_z}{\partial y^2}\right) = 0\\
    \frac{\partial B_x}{\partial x} + \frac{\partial B_y}{\partial y} = 0
\end{align*}
\end{subequations}

The values for $\nu$ and $\eta$ are set by hyperparameter optimization with respect to the overall training loss. The final values suggested by the optimization are: $\nu = 1.24 * 10^{-4}$ and $\eta = 1.88 * 10^{-3}$ (GEM); $\nu = 1.58 * 10^{-3}$ and $\eta = 6.87 * 10^{-3}$ (LW3); $\nu = 2.76 * 10^{-3}$ and $\eta = 5.59 * 10^{-3}$ (OT). We note that [\cite{bard21}] tested the effect of changing $\nu$ on the reconstructed solution for a 1D MHD shock tube; that study found that the best value for $\nu$ depends on the problem. This motivated us to use hyperparameter optimization for choosing appropriate values.

\end{document}